\pdfoutput=1
\documentclass[11pt]{article}

\usepackage[preprint]{acl}

\usepackage{times}
\usepackage{latexsym}
\usepackage{amsmath}

\usepackage[T1]{fontenc}
\usepackage[utf8]{inputenc}
\usepackage{xcolor}
\usepackage{soul}
\usepackage{microtype}

\usepackage{inconsolata}

\usepackage{graphicx}
\usepackage{multirow}
\title{Personality Prediction from Life Stories using Language Models}

\author{%
\small            % compact the whole header
\centering        % centre the entire block on the page
\begin{tabular}{ccc}
\textbf{Rasiq Hussain} & \textbf{Jerry Ma} & \textbf{Ritik Khandelwal} \\[2pt]
\textnormal{Southern Methodist University}& \textnormal{Southern Methodist University}& \textnormal{Southern Methodist University}\\[2pt]
\texttt{rasiqh@mail.smu.edu} & \texttt{jerryma@mail.smu.edu} & \texttt{rkhandelwal@mail.smu.edu} \\[12pt]
\multicolumn{3}{c}{%
  \begin{tabular}{cc}
  \textbf{Joshua Oltmanns} & \textbf{Mehak Gupta} \\[2pt]
  \textnormal{Washington University in St. Louis}& \textnormal{Southern Methodist University}\\[2pt]
  \texttt{j.oltmanns@wustl.edu} & \texttt{mehakg@mail.smu.edu}
  \end{tabular}
}
\end{tabular}
}

\begin{document}
\maketitle
\begin{abstract}
Natural Language Processing (NLP) offers new avenues for personality assessment by leveraging rich, open-ended text, moving beyond traditional questionnaires. In this study, we address the challenge of modeling long narrative interview where each exceeds 2000 tokens so as to predict Five-Factor Model (FFM) personality traits. We propose a two-step approach: first, we extract contextual embeddings using sliding-window fine-tuning of pretrained language models; then, we apply Recurrent Neural Networks (RNNs) with attention mechanisms to integrate long-range dependencies and enhance interpretability. This hybrid method effectively bridges the strengths of pretrained transformers and sequence modeling to handle long-context data. Through ablation studies and comparisons with state-of-the-art long-context models such as LLaMA and Longformer, we demonstrate improvements in prediction accuracy, efficiency, and interpretability. Our results highlight the potential of combining language-based features with long-context modeling to advance personality assessment from life narratives.

%provide superior performance in accuracy, efficiency, and interpretability by using meaningful contextual text embeddings from pre-trained models, while capturing full-context with RNN attention layers effectively manage the long context and provide interpretability — a limitation of LLMs. We compared with state-of-the-art LLMs with extended token limits, such as LLaMA and Longformer. While these models can process longer texts, our proposed hierarchical method demonstrated superior performance in accuracy, efficiency, and interpretability. 

%reduce the residual error from the pre-trained model  use lengthy life narrative texts from older adults, each exceeding 2000 tokens to . We employ a hierarchical transformer approach with stop-gradient technique, involving the fine-tuning of pre-trained language models with a sliding-window technique, followed by the use of Recurrent Neural Networks (RNNs) with attention layers to reduce the error from the pre-trained model and capture the full context. 
\end{abstract}

\section{Introduction}

Self-report measures have long been the standard in psychological assessment due to their simplicity and ability to generate quantifiable personality scores \citep{piotrowski1999assessment,soto2019replicable}. However, they rely on self-awareness, are prone to response biases, and are underutilized in clinical practice since only about 50\% of psychologists report using standardized assessments regularly \citep{wright2017assessment}. Many practitioners have shown growing interest in integrating richer modalities, such as language-based tools, into assessment frameworks \citep{paulhus2007self,hatfield2004use,wright2017assessment,kjell2022natural}.

As a result, there is growing momentum toward language-based personality assessment, where individuals own narratives are used to derive personality scores. Advances in Natural Language Processing (NLP) now make it feasible to analyze free-form text and extract trait-relevant information in a systematic and interpretable way.

In this work, we focus on predicting Five-Factor Model (FFM) personality scores from transcribed life narrative interviews, each averaging over 2000 tokens. Existing pre-trained language models (PLMs) such as BERT and RoBERTa, which are limited to processing 512 tokens at a time cannot contextualize long interviews without adaptations. Although newer LLMs can technically scale to longer contexts, real-world constraints (memory, training time, architecture stability) limit their effectiveness. Furthermore, their black-box nature limits interpretability, an important factor for clinical applications where understanding model reasoning is crucial.

To address these limitations, we propose a lightweight, interpretable two-step architecture using small PLMs and RNNs \citep{pappagari2019hierarchical}. Our method is designed to handle long text efficiently while producing meaningful and transparent personality predictions. %Our aim is to build an efficient and less computationally expensive method to handle long inputs and provide interparatibility to otherwise black-box models. 
The main contributions of this paper can be listed as follows:
\begin{enumerate}
    \item We propose a 2-step method with a stop-gradient technique; first fine-tune RoBERTa with a sliding window, and freezing its weights; second, using an RNN on the sequential window embeddings to reduce residual error of RoBERTa and capture full-context of long text to predict personality scores. 
    \item We analyze attention weights from RNN attention layers through different interpretability techniques to validate the model’s results with the help of domain experts. 
    \item We conduct ablation studies and compare our approach with fine-tuned baselines (RoBERTa, XLNet, Longformer, LLaMA), demonstrating the effectiveness of our approach in terms of performance, efficiency and interpretability. 
\end{enumerate}

\section{Related Work}

The link between language and psychological phenomena has long been established \citep{boyd2021natural,tausczik2010psychological}. NLP and AI tools have been applied to understand identity \citep{argamon2007mining,berger2022using,kwantes2016assessing,schwartz2013personality}, emotion \citep{de2022joint,eichstaedt2018facebook,sun2020language}, behavior \citep{curtis2018can,kjell2021computational,macavaney2021community}, and other psychological traits \citep{eichstaedt2021closed,iliev2015automated,jackson2022text,schwartz2015data}. Within personality research, the Five Factor Model (FFM) has been shown to be detectable through language \citep{park2015automatic}.

Recent NLP advances, particularly deep learning and large language models (LLMs), have improved trait detection. Models like BERT and ChatGPT have achieved high performance, mostly for trait classification \citep{jain2022personality,mehta2020bottom,ganesan2023systematic,peters2024large}. However, personality is best modeled as continuous dimensions. To this end, several studies have used fine-tuned transformers on social media text to predict continuous FFM scores\citep{cutler2023deep,matero2021evaluating,liu2019roberta,simchon2023online,ganesan2023zeroshot}.

Social media, with its short-length text, fits well within the 512-token limits of standard transformers \citep{liao2021improved,joshy2022analyzing}.
In contrast, this work uses transcribed life narratives which are spoken interviews averaging over 2000 tokens to predict personality traits. Long-form texts provide richer information but pose challenges for standard models. Techniques like sliding windows can split text into overlapping segments \citep{wang2019multi}, but this often loses global context and leads to poorer performance.

LLMs like Llama and Longformer provide extended token limits, but are highly expensive and slow to train \citep{zhang2020bert,ding2020cogltx}. Moreover, LLMs with extended token limits still struggle to fully utilize information within the long context, known as the lost-in-the-middle challenge. \citet{an2024make} and \citet{geng2024breaking} addresses this issue by selecting a few important and unique segments, respectively. %\citet{geng2024breaking} reduces computational redundancy in encodings by saving unique user behavior embeddings. 
Transformer-XL \citep{dai2019attentive} and  Compressive Transformer \citep{rae2019compressive}  adds memory cells and condense history information respectively. Memformer \citep{wu2020memformer} retrieve and update the memory dynamically and ERNIE-Doc \citep{ding2020ernie} concatenates the history segment to the current segment.

To efficiently capture long-text context using models with fewer parameters and lower computational cost, \citet{zhang2020bert} proposed BERT-AL, combining BERT with multi-channel LSTMs to summarize long text.
\citet{geng2024breaking} handles long sequences by aggregating embeddings of smaller tokens. \citet{lynn2020hierarchical} used a two-level RNN attention model, one for word and another for message-level which were on concatenated Facebook posts to predict personality. In contrast, we apply RNN attention only at the segment level, where segments are overlapping sliding windows from long interview transcripts. Unlike \citet{lynn2020hierarchical}, we first fine-tune a language model to extract contextual embeddings for each segment before feeding them into the RNN. Hierarchical approaches like those in \citet{pappagari2019hierarchical,gong2020recurrent} combine transformer and recurrent layers over segment embeddings. We adopt a similar strategy but enhance it with attention layers for interpretability. Unlike prior work, our two-step training with fine-tuning and freezing the language model before training the RNN improves gradient stability and efficiency, especially with PLMs. It also supports more stable segment-level interpretability via RNN attention weights \citep{li2015visualizing}.

%By using the hierarchical architectures \citep{pappagari2019hierarchical,gong2020recurrent} we will combine the pre-trained language model with RNN to take advantage of contextual embeddings from pre-trained language models and capture the full context of the lengthy interview with RNN. We will train this model for a continuous approach to personality prediction that aligns with the dimensional perspectives advocated by researchers like \citep{widiger2005dimensional}.%to predict the continuous FFM personality scores. We will also implement an attention layer in RNN to introduce a comprehensive interpretability regime to prove our model's correctness. 

\section{Data}

%\subsection{Data Collection}
%\citep{oltmanns2014prevalence}

The dataset used in this study is derived from the \textit{St. Louis Personality and Aging Network}  and comprises life narrative interviews with 1,408 older adults aged 55 to 64 from the St. Louis area. Participants recount their lives in 4 chapters starting from 18 years of age, describing key events and characters, and their recorded language is used to predict FFM personality traits. This approach facilitates a deep understanding of participant's life experiences and closely aligns with understanding personality through life stories, as described by McAdams \citep{mcadams1993stories}. %The life narrative interview was adapted from \citet{mcadams1993interviewer}, where participants were asked to share their life stories in 3 to 4 chapters, starting from age 18. They also detailed significant characters, high and low points, and major turning points in their lives. 
Participants also completed the NEO-Personality Inventory-Revised (NEO-PI-R; \citet{costa2008revised}), which includes 240 items rated from 0 (strongly disagree) to 4 (strongly agree) that score five trait domains: openness, conscientiousness (or constraint v/s disinhibition), extraversion (v/s introversion),  agreeableness (v/s antagonism), and neuroticism (or emotional instability v/s stability) \citep{Widiger2019}. %Transcripts were deidentified using named entity recognition.  %Additionally, socioeconomic data, such as household income, parental education, and participant education, were collected through a demographic questionnaire to supplement the qualitative interview data.

%The life narrative transcripts collected, utilize the Five Factor Model (FFM) personality traits for assessment. The FFM of general personality structure consists of the five broad domains of personality traits: openness (or unconventionally), conscientiousness (or constraint vs. disinhibition), extraversion (vs. introversion),  agreeableness (vs. antagonism), and neuroticism (or emotional instability vs. stability) \citep{Widiger2019}. 

%The raw text transcripts obtained from the interviews underwent preprocessing to prepare the data for training transformer-based models. %To prepare the interviews for preprocessing, raw text files were extracted from the original dataset and their corresponding interview IDs, which served as unique identifiers for each interview. 
%The data was formatted into JSON files appropriate for training, where each entry included the complete text and corresponding FFM scores to evaluate five domain personality traits. 
The interviews were processed to remove any commentary on the audio transcripts and to distinguish between the interviewer's and the interviewee's speech. On average each transcript contained 2,513 words. Out of 1417 transcripts, we excluded 8 transcripts that were less than 50 words. We provide more details about transcript lengths and score range for each personality trait in Table \ref{tab:data}.

\begin{table}[!t]
\small
\centering
\begin{tabular}{|l|l|l|l|l|}
\hline
 & \textbf{Min}        & \textbf{Max}  & \textbf{Mean} & \textbf{Std. Dev}\\\hline 

Length    &151&19,020&  2,513&      1,910\\  \hline

O&51&211.5&    113.84&    19.31\\  \hline
C&39&241.5&     125.19&   19.7\\  \hline
E&47&216&     109&   19.4\\  \hline
A&66&220.5&    131.45&    16.5\\  \hline
N&13&162&    72.97&    21.7\\  \hline
\end{tabular}
\caption{Data Statistics about life narrative interviews. It shows the length of transcripts in words and kb. Last 5 rows show the range of each personality score. }\label{tab:data}
\end{table}

%All text for 1409 transcripts was converted to lowercase, tokenized, and stop-words were removed using the Natural Language Toolkit (NLTK)\citep{loper-bird-2002-nltk}. The cleaned text enabled clearer attribution of dialogue and enhanced readability for subsequent stages of analysis. For tokenization, we employed AutoTokenizer from the Hugging Face library to tokenize the text data. 

\section{Models and Methods}\label{sec:4}
Below we describe our strategies for handling long life narrative interviews using both small and large language models, followed by our proposed two-step approach for efficient, interpretable modeling.

\subsection{Fine-tuning smaller language models}\label{sec:small}

We fine-tuned large versions of RoBERTa (355M), XLM-RoBERTa (550M), and XLNet (340M) to predict the five personality traits from interview transcripts. While these models are effective for contextual language tasks, their 512-token limit poses a challenge given our transcripts average ~2992 tokens.

\paragraph{Sliding-window for long text:}
To address the token limit, we applied a sliding-window approach, segmenting each transcript into overlapping 512-token chunks. We fine-tuned each model on these segments and took the median prediction across windows to produce final personality scores (see Supplementary \ref{apx:fine} for details).

While this allows long text processing with smaller PLMs, it treats windows independently and fails to capture broader narrative context. To address this, we explore: (1) large language models with extended token limits, and (2) sequential aggregation of window embeddings obtained from smaller PLMs using a lightweight model—our proposed approach offering a more computationally efficient alternative to LLMs.
%The sliding-window technique enables the processing of long texts using language models with limited token capacities. However, this approach has a key limitation: each prediction is made independently for each window, without accounting for information in other segments, thereby missing the full context of the life narrative. This limitation can be mitigated in two ways: (1) by leveraging large language models (LLMs) with extended token limits, or (2) by aggregating window-level embeddings from a fine-tuned smaller language model using an additional sequence model to capture the broader context. In the next two sections, we first discuss the use of LLMs for handling long text, followed by our proposed approach that applies sequential modeling over contextual embeddings from a smaller language model, offering a more computationally efficient alternative to LLMs.

\subsection{Fine-tuning large language models}\label{sec:large}
To process long life narrative interviews in a single window, we fine-tuned both Longformer and LLaMA models.
Longformer, with 435 million parameters, uses sparse attention to reduce complexity and process up to 4,096 tokens efficiently without overlapping windows. However, its sparse attention limits the capture of long-range dependencies, making it less effective at retaining full-context beyond its attention window.

 \begin{figure}[!b]

     \centering
     \includegraphics[width=0.95\linewidth]{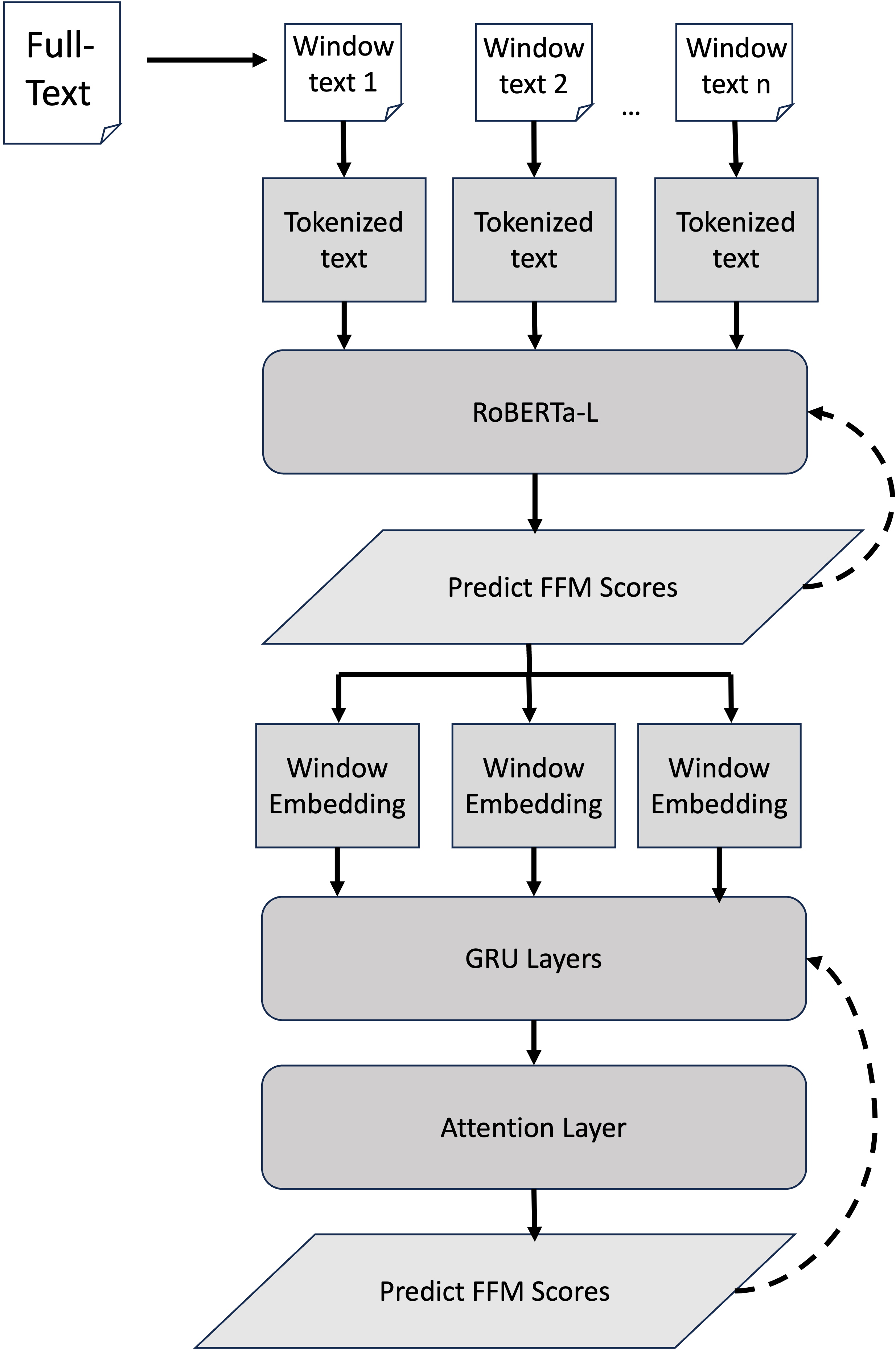}
     \caption{RoBERTa+RNN architecture. Dashed arrows show backpropagation with 1) fine-tuning RoBERTa with sliing-windows on FFM scores and 2) training RNN with embeddings from fine-tuned RoBERTa.}
     \label{fig:model}
 \end{figure}

LLaMA, a large autoregressive model, was adapted for regression by adding a two-layer MLP head. We used LLaMA 3.1 base (foundational) model, which is capable of handling sequences up to 128K tokens in a single pass but has quadratic increases in computational and memory demands with sequence length. To reduce the significant computational demands of full fine-tuning, we applied parameter-efficient fine-tuning (PEFT) using the LoRA method via the SFTTrainer. We configured LoRA with a rank 
$r$=64, a scaling factor $lora_{\alpha}$
 =16, and a dropout of 0.1. See Supplementary \ref{apx:llama} for full hyperparameters.

Despite their ability to process full narratives, Longformer’s sparse attention and LLaMA’s memory demands make them less practical for scalable clinical use compared to compact models with contextual aggregation, as introduced in our two-step method discussed in the next section.

\subsection{Language model and Recurrent Neural Network}\label{sec:prop}

%To capture full-context from long text efficiently, we combine smaller PLMs with RNNs. This is done using a two-step approach: first, RoBERTa is fine-tuned via a sliding-window strategy to predict FFM scores.

To efficiently capture the full-context of long text while taking advantage of contextual language understanding of PLMs without increasing computational and memory cost we combine smaller PLMs (RoBERTa) with RNNs. This is done using a two-step approach: first, RoBERTa is fine-tuned via a sliding-window strategy to predict FFM scores. The fine-tuned model is then frozen and used to extract [CLS] embeddings from each window. Given input transcripts \( X = \{X_1, X_2, \dots, X_N\} \), we apply a sliding window of size \( w \) and stride \( s \). For each transcript \( X_i = \{x_{t_i}, x_{t_i+1}, \dots, x_{t_i+w-1}\} \), we pass its windows through a transformer model \( \mathcal{T}(\cdot) \) which is RoBERTa-L in our implementation and extract the \texttt{[CLS]} token embedding of 1024-dimension:

\begin{equation}
(CLS_1,\dots,CLS_T) = \mathcal{T}(X_i)_{\texttt{[CLS]}} 
\end{equation}
This model \( \mathcal{T}(\cdot) \) is trained on MSE loss $L_{{\mathcal{T}}_{MSE}}$ to predict personality scores. These embeddings $(CLS_1,\dots,CLS_T)$ are sequenced and passed to an RNN with attention to predict FFM scores, where each transcript yields up to 200 embeddings and shorter sequences are padded. We used two layers of GRU with a hidden size of 256 for RNN model. During RNN training, gradients flow only through the RNN, reducing residual error from RoBERTa. 

\begin{equation}
 \{h_1, \dots, h_T\} = \text{RNN}(\text{CLS}_1, \dots, \text{CLS}_T)
\end{equation}

\begin{equation}
 \alpha_t = \frac{\exp(\mathbf{w}^\top h_t)}{\sum_{t=1}^T \exp(\mathbf{w}^\top h_t)},  c_i = \sum_{t=1}^T \alpha_t \cdot \text{h}_t
 \end{equation}
%  \begin{equation}
%  c_i = \sum_{t=1}^T \alpha_t \cdot \text{h}_t
% \end{equation}

where $c_i$ is the context vector for transcript $X_i$ obtained after weighted sum of hidden states corresponding to $CLS$ vector of each window $x_t$. RNN model is separately trained on MSE loss $L_{{RNN}_{MSE}}$ to predict personality scores. 
Figure \ref{fig:model} illustrates the architecture. 

%We train this model in the 2-step method, where we first fine-tune the RoBERTa model using the sliding-window technique to predict FFM scores. The fine-tuned RoBERTa model is then frozen and used to extract the CLS embeddings of each sliding window from its last layer. We arrange these embeddings longitudinally and train RNNs with attention mechanisms to predict the FFM scores from life narrative texts. In the RNN training step, the gradient only flows through the RNN model to reduce the residual error of RoBERTa fine-tuning. The model architecture and training strategy are shown in Figure \ref{fig:model}. 

%RNNs effectively model sequential language dependencies \citep{Shiri2023ACO}. We use Gated Recurrent Units (GRU) \citep{Shiri2023ACO}, a variant that mitigates vanishing gradients and merges input and forget gates into a single update gate, improving training speed and efficiency.

\begin{table*}[t]
  \small
\centering
  
  \begin{tabular}{c|c|c|c|c|c|c|c|c|c|c|}
   \multirow{2}{*}{Model}& \multicolumn{2}{c|}{O} &  \multicolumn{2}{c|}{C} &  \multicolumn{2}{c|}{E} &  \multicolumn{2}{c|}{A} &  \multicolumn{2}{c|}{N}\\
  &MSE&$R^2$&MSE&$R^2$&MSE&$R^2$&MSE&$R^2$&MSE&$R^2$\\\hline
  
  %Roberta&0.82&0.17&0.95&0.04&0.85&0.14&0.93&0.07&0.89&0.10\\
  
   % \multicolumn{11}{c}{Small Language Models}\\\hline
  \multirow{2}{*}{Roberta-L}&0.81&0.19&0.90&0.10&0.88&0.12&0.96&0.04&0.81&0.19\\
  &(0.01)&(0.01)&(0.05)&(0.05)&(0.03)&(0.03)&(0.03)&(0.03)&(0.04)&(0.04)\\\hline
  
  XLM-&0.81&0.19&0.97&0.03&0.88&0.12&0.93&0.07&0.90&0.10\\
  RoBERTa&(0.02)&(0.02)&(0.04)&(0.05)&(0.03)&(0.04)&(0.02)&(0.02)&(0.04)&(0.04)\\\hline
  \multirow{2}{*}{XLNet-L}&0.83&0.17&0.99&0.01&0.90&0.10&0.89&0.03&0.81&0.19\\
  &(0.01)&(0.01)&(0.05)&(0.05)&(0.04)&(0.04)&(0.03)&(0.02)&(0.04)&(0.03)\\\hline 
 %Roberta+NN & 0.23 &0.76  & 0.23 &0.76& 0.22 & 0.77&0.25  & 0.73 &0.27  &0.72   \\
% &(0.01)&(0.01)&(0.05)&(0.05)&(0.03)&(0.03)&(0.02)&(0.02)&(0.04)&(0.04)\\
 %  \multicolumn{11}{c}{Large Language Models}\\\hline
\multirow{2}{*}{Longformer}&0.82&0.18&0.97&0.03&0.91&0.09&0.96&0.04&0.91&0.10\\
  &(0.03)&(0.03)&(0.05)&(0.05)&(0.04)&(0.04)&(0.04)&(0.04)&(0.04)&(0.04)\\\hline

  %Roberta+RNN &0.22&0.77&0.2&0.78&0.22&0.77&0.24&0.75&0.21&0.77\\
  %w/o Attention&&&&&&&&&&\\

% Llama3.1&1.12&-0.11&1.27&-0.27&1.16&-0.16&1.16&-0.16&1.15&-0.15\\
%   1B&(0.02)&(0.02)&(0.04)&(0.04)&(0.03)&(0.03)&(0.03)&(0.03)&(0.04)&(0.04)\\\hline
%   Llama3.1&1.33&-0.33&1.39&-0.39&1.42&-0.42&1.10&-0.10&1.32&-0.32\\
% 8B&(0.01)&(0.01)&(0.02)&(0.02)&(0.03)&(0.03)&(0.02)&(0.02)&(0.03)&(0.03)\\\hline
 Llama3.1&1.40&-0.40&1.19&-0.19&1.12&-0.12&1.33&-0.33&1.21&-0.21\\
8B&(0.02)&(0.01)&(0.02)&(0.02)&(0.02)&(0.02)&(0.01)&(0.01)&(0.02)&(0.03)\\\hline
 %     FTemb+RNN  &0.69&0.14 &1.0&0.05&0.7&0.15&0.8& 0.08&0.79&0.12\\
 % with Attention&(0.01)&(0.01)&(0.04)&(0.04)&(0.04)&(0.03)&(0.02)&(0.01)&(0.02)&(0.01)\\

 FT-RobertaL  &\textbf{0.69}&\textbf{0.30} &\textbf{0.68}&\textbf{0.30}&\textbf{0.46}&\textbf{0.50} &\textbf{0.56}& \textbf{0.43}&\textbf{0.47}&\textbf{0.52}\\
 +RNN&(0.01)&(0.01)&(0.05)&(0.05)&(0.03)&(0.03)&(0.02)&(0.02)&(0.02)&(0.02)
 \\

  \end{tabular}
  \caption{Comparison of mean MSE and $R^2$ with std dev over 5-fold cross-validation between all baselines and our proposed model. Fine-tuned (FT).}\label{tab:results}
\end{table*}

 \begin{table*}[t]
 \small
\centering
  
  \begin{tabular}{c|c|c|c|c|c|c|c|c|c|c|}
   \multirow{2}{*}{Model}& \multicolumn{2}{c|}{O} &  \multicolumn{2}{c|}{C} &  \multicolumn{2}{c|}{E} &  \multicolumn{2}{c|}{A} &  \multicolumn{2}{c|}{N}\\
  &MSE&$R^2$&MSE&$R^2$&MSE&$R^2$&MSE&$R^2$&MSE&$R^2$\\\hline
  
  %Roberta&0.82&0.17&0.95&0.04&0.85&0.14&0.93&0.07&0.89&0.10\\

 FT-RobertaL  &0.86&0.03&0.99&0.007&0.5&0.02&0.9&0.05&1.1&0.01\\
 +RNN CF&0.02)&(0.01)&(0.05)&(0.03)&(0.03)&(0.02)&(0.01)&(0.01)&(0.03)&(0.03)
 \\\hline
 FT-RobertaL  &1.16&-0.30&0.9&-0.01&0.96&-0.03&1&0.007&0.9&0.01\\
 +TF CF&0.02)&(0.03)&(0.05)&(0.04)&(0.03)&(0.03)&(0.02)&(0.01)&(0.03)&(0.04)
 \\\hline
  FT-RobertaL  &0.79&0.14&0.89&0.05&0.84&0.15&0.89&0.08&0.81&0.15\\
 +FFN 2-step&(0.01)&(0.01)&(0.05)&(0.05)&(0.03)&(0.03)&(0.02)&(0.02)&(0.02)&(0.02)
 \\\hline
  
 FT-RobertaL  &0.75&0.20 &0.9&0.1&0.70&0.20&0.80&0.1&0.79&0.18\\
 +TF 2-step&(0.02)&(0.02)&(0.03)&(0.04)&(0.02)&(0.02)&(0.02)&(0.02)&(0.03)&(0.02)
 \\\hline

 PT-RobertaL  &0.86&0.13&1.01&0.02&0.90&0.10&1.12& 0.02&0.90&0.10\\
 + RNN 2-step&(0.01)&(0.01)&(0.04)&(0.04)&(0.04)&(0.03)&(0.02)&(0.01)&(0.02)&(0.01)\\\hline
 
FT-RobertaL  &\textbf{0.69}&\textbf{0.30} &\textbf{0.68}&\textbf{0.30}&\textbf{0.46}&\textbf{0.50} &\textbf{0.56}& \textbf{0.43}&\textbf{0.47}&\textbf{0.52}\\
 +RNN 2-step&(0.01)&(0.01)&(0.05)&(0.05)&(0.03)&(0.03)&(0.02)&(0.02)&(0.02)&(0.02)
 \\

  \end{tabular}
  \caption{Comparison of mean MSE and $R^2$ with std dev over 5-fold cross-validation between all model variations. Fine-tuned (FT), Pre-trained (PT), Continuous gradient flow (CF), Transformer (TF), Feed-forward Network (FFN).}\label{tab:variations}
\end{table*}
 
%We used two layers of GRU with a hidden size of 256 each. %Each input is a vector of 1024-dimensional [CLS] embeddings (one per sliding window) capturing the full-context of transcripts. Each transcript yields up to 200 embeddings and shorter sequences are padded.

%a sequence of embeddings for each sliding-window for each participant is given to the GRU layers as input. The embeddings used to train the RNN consisted of a sequence of vectorized representations where each embedding is a fixed-size vector of 1024 dimensions that encodes information from each sliding-window of the transcript. The maximum sequence length (number of windows) for each sample is 200 and padding was applied to samples of lesser length.

Attention layers enhance RNN interpretability by assigning scores to GRU outputs, highlighting key text segments. A softmax function normalizes these scores, producing a weighted sum used to generate a context vector. This vector is fed into a fully connected layer to predict FFM scores, enabling long-term context modeling with focused attention on important text segments.

\section{Experiments and Results}
All models were trained using 5-fold cross-validation to ensure robust evaluation across data subsets. An 80:20 train:test split was applied and kept consistent across both stages of the two-step method. Additionally, 5\% of the training set was used for validation during hyperparameter tuning. The best-performing model after tuning was selected for final evaluation. Performance was assessed using mean squared error (MSE) and R² score. See Supplementary \ref{apx:fine}, \ref{apx:llama}, and \ref{apx:hyp} for hyperparameter details. %MSE provided insight into the presence of larger prediction errors by squaring the residuals and $R^2$ score was used to assess the proportion of variance explained by the model, with higher values indicating a better fit.  %We report average $R^2$ and MSE scores across 5-folds .

\subsection{Comparison Experiments}
We conduct comprehensive experiments comparing our proposed model against five existing language models, as well as five ablations of our own architecture. These comparisons are designed to assess the impact of fine-tuning, architectural choices, and training strategies on model performance.
\paragraph{Baseline Models:} For baseline comparison, we compare our model proposed in Section \ref{sec:prop} with existing smaller PLMs discussed in Section \ref{sec:small} and LLMs discussed in Section \ref{sec:large}.

\paragraph{Continuous Flow (CF) vs. 2-Step Training:} To evaluate the effectiveness of our two-step training strategy, we compare it with a continuous flow (CF) setup, where gradients from the downstream RNN are backpropagated into RoBERTa during joint training. We also evaluate a variant of CF by replacing the RNN with a Transformer-based encoder (TF) to assess the impact of different sequence models in an end-to-end framework.

\paragraph{Ablation Studies on Sequence Modeling and Fine-Tuning:} To analyze the effect of the sequence modeling component in our two-step framework, we replace the RNN with (1) a feed-forward network (FFN) and (2) a Transformer (TF). FFN model takes mean CLS embeddings over all windows whereas transformer takes all window embeddings with segment embeddings to maintain text sequence. Additionally, we evaluate the importance of fine-tuning by comparing our fine-tuned RoBERTa+RNN model with a variant that uses frozen (non-fine-tuned) RoBERTa embeddings.

\subsection{Results}
\paragraph{Baseline Comparison:}

%\subsection{Comparative Results}
Table \ref{tab:results} compares $R^2$ and MSE scores between smaller PLMs, LLMs and our proposed method. These models are fine-tuned on our dataset to predict personality scores. %We also compare our fine-tuned RoBERTa-L+RNN approach with another model where we extract embeddings from pre-trained models and use those to train RNN model for FFM scores. 
Results show that finetuned RoBERTa + RNN trained with our proposed 2-step method provides the minimum MSE and highest $R^2$ across all personality traits. 

\begin{table}[!b]
\small
\centering
\begin{tabular}{|l|c|c|c|c|}
\hline
\textbf{Model}  & \textbf{\#Params}        & \textbf{Train}  & \textbf{GPU} & \textbf{TFLOPs}\\ 
\textbf{}   &              & \textbf{Time}  & \textbf{(GB)}&\\\hline
Roberta-L     &335 mil& 5:42& 13& 2.2\\  \hline

XLM        &550 mil & 6:28&20& 3.4\\ 
RoBERTa-L        &&&   &\\ \hline
XLNet-L           &     340 mil     & 5:40&15&  2.1\\ \hline
Longformer         &    435 mil    & 8:04&17& 1.8\\  \hline
% Llama-1B        &      1 billion        & Vey High                    & 20-22           \\ \hline
Llama-8B         &     8 billion         & 2:36&70& 82\\ \hline
\end{tabular}
\caption{Comparison of training times, model sizes and TFLOPs for different fine-tuned models.}\label{tab:resource}
\end{table}

\paragraph{Ablation Analysis:}  We compare modeling strategies to assess the impact of fine-tuning and training design. As shown in Table \ref{tab:variations}, continuous flow (CF) variants consistently underperform, suggesting that jointly training RoBERTa with the downstream model may hinder convergence. Among the sequence models, RNN outperforms both Transformer (TF) and feedforward network (FFN), highlighting the advantage of capturing temporal dependencies in life narratives. FFN performs the worst, indicating that ignoring sequential structure limits predictive accuracy. Also, TF offer no clear advantage in our setting with manageable sequence lengths ($<=$ 200), while RNNs provide a better balance of accuracy, interpretability, and efficiency. Using pre-trained RoBERTa without task-specific fine-tuning also degrades performance, reinforcing the importance of domain adaptation. Our proposed two-step model achieves the highest $R^2$ across all traits, outperforming all ablations.

\begin{figure}[t]
\centering
\includegraphics[width=0.39\textwidth]{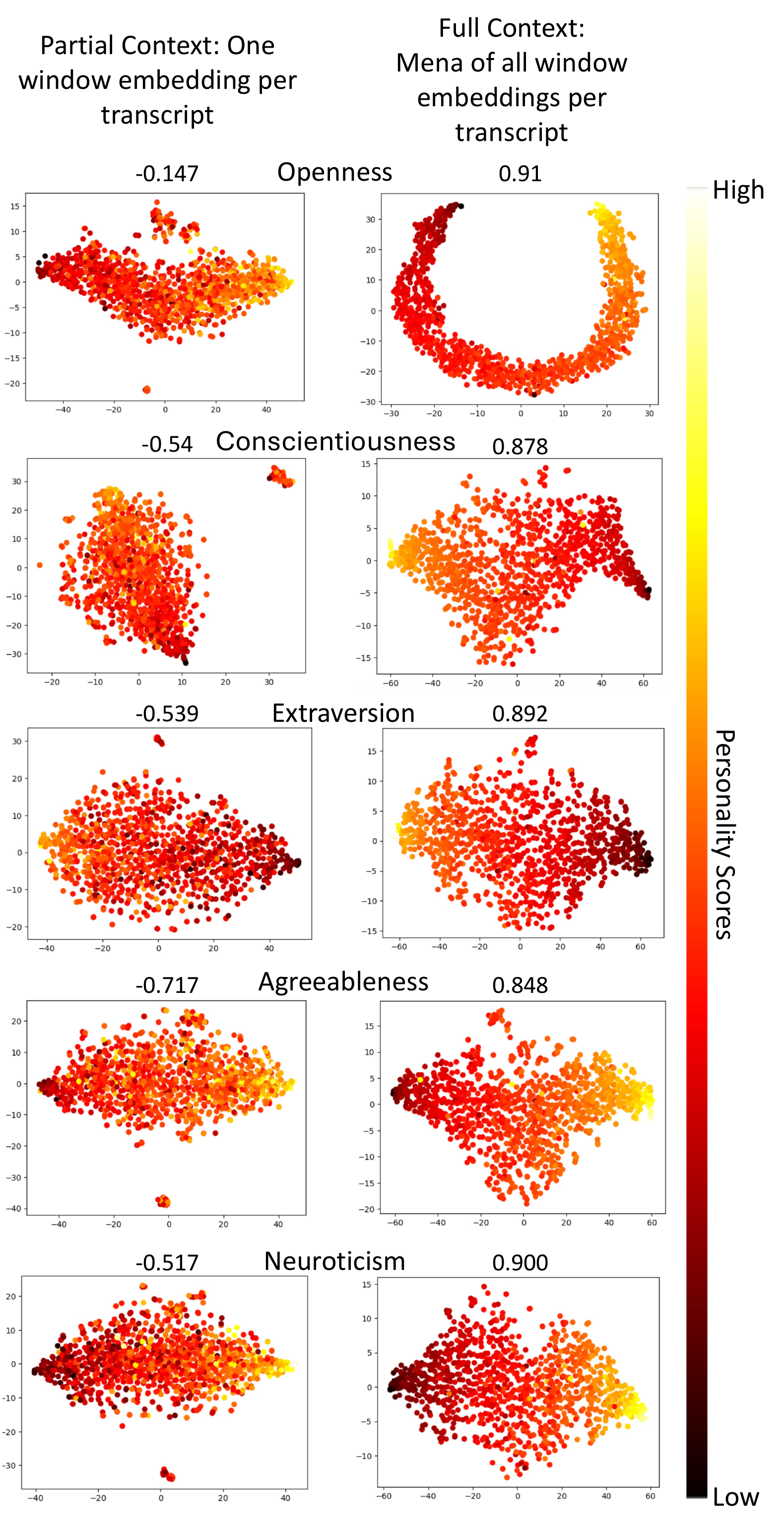}
%\vspace{-5pt} % Reduce space between the figure and caption
\caption{2-D visualization of [CLS] embeddings of training set from last layer of fine-tuned RoBERTa. The left panel shows embedding from a randomly chosen window per transcript, and the right panel shows the mean of all window-level embeddings. Ridge regression (where embeddings are used to predict personality scores) $R^2$ values are displayed at the top of each image.}
\label{fig:emb}
\end{figure}

\paragraph{Comparison of Time and Resource efficiency} 

We evaluate memory and computational efficiency of baseline models in Table \ref{tab:resource}, focusing on parameter count, training time, GPU memory usage, and estimated FLOPs per sample. All models were fine-tuned with batch size 8 on NVIDIA A100 GPU for fair comparison. Among models with the standard 512-token limit, RoBERTa-L is the most efficient, offering low training time and FLOPs. While Longformer and LLaMA-8B handle longer inputs, they require more memory and training resources. LLaMA-8B, with 8 billion parameters, is especially costly—requiring an estimated 82 TFLOPs per sample, nearly 40× that of RoBERTa-L.

 \begin{figure*}[t]
     \centering
     \includegraphics[width=1\linewidth]{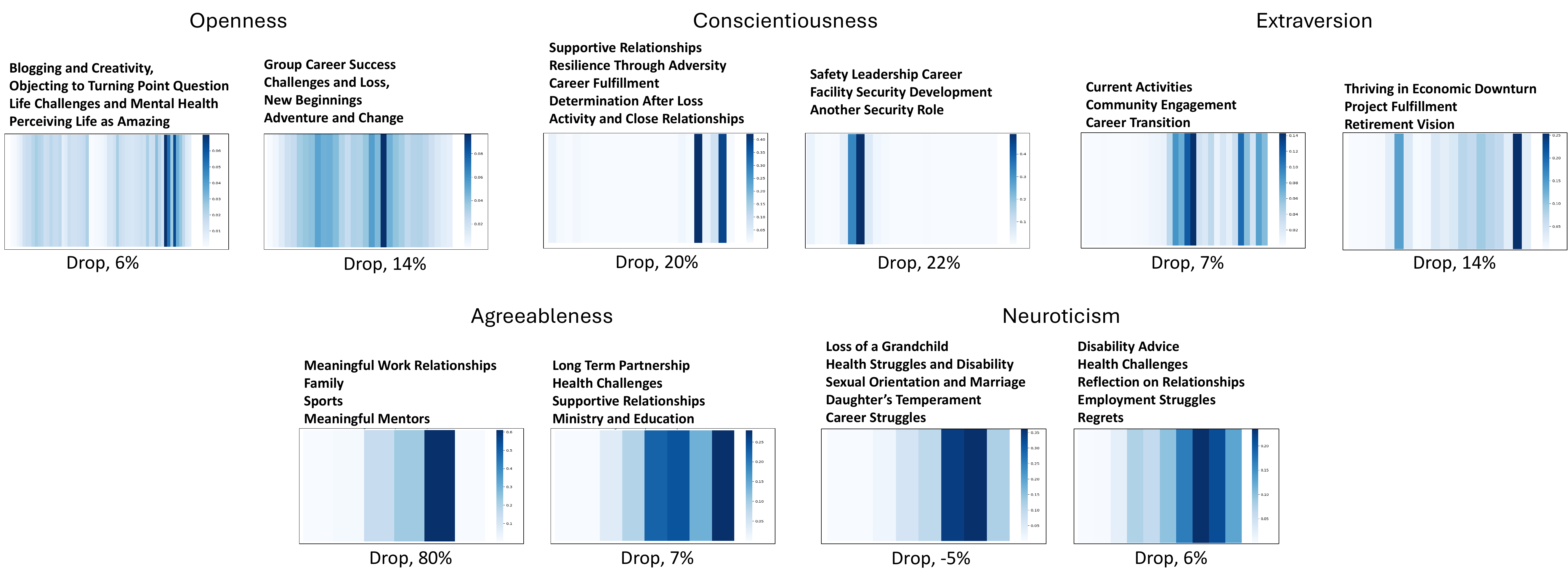}
     \caption{It shows attention given to different text windows in the transcript for two random examples from the top 10 predicted personality scores for each trait from the test dataset. The text at the top shows the list of topics discussed in the 3 highest attention windows. The \%age at the bottom shows the drop in the predicted score if the text from the one highest windows is removed.}
     \label{fig:heat}
 \end{figure*}
\subsection{Embedding Visualization}
To investigate the advantage of capturing full-context versus partial-context from life narrative interviews, we visualize high-dimensional text embeddings using t-SNE, derived from the training dataset. In Figure~\ref{fig:emb}, we show two visualizations: the [CLS] embeddings from a randomly selected window (left panel) and the mean [CLS] embeddings across all windows for a participant (right panel), both from the final layer of fine-tuned RoBERTa. These embeddings are projected into 2D space and displayed as heatmaps, where color intensity corresponds to the ground-truth score shown by the colorbar on right.

The left panel, representing partial context from a single window, shows no clear separation between high and low personality scores, indicating insufficient information for accurate prediction. In contrast, the right panel reveals a smoother transition from dark to light, suggesting that aggregating window-level embeddings provides better context. We further support this observation with ridge regression $R^2$ scores displayed on top of each embedding, demonstrating the benefit of full-context embeddings.

\section{Model Interpretability }

One of the key aspects of this study was about the RNN model's interpretability. We have validated high-attention text segments with domain experts to ensure reliable predictions. These experiments provide insights into the model's decision-making and support its accuracy.

\subsection{RNN Attention Interpretability}

 \begin{figure}[t]
     \centering
     \includegraphics[scale=0.15]{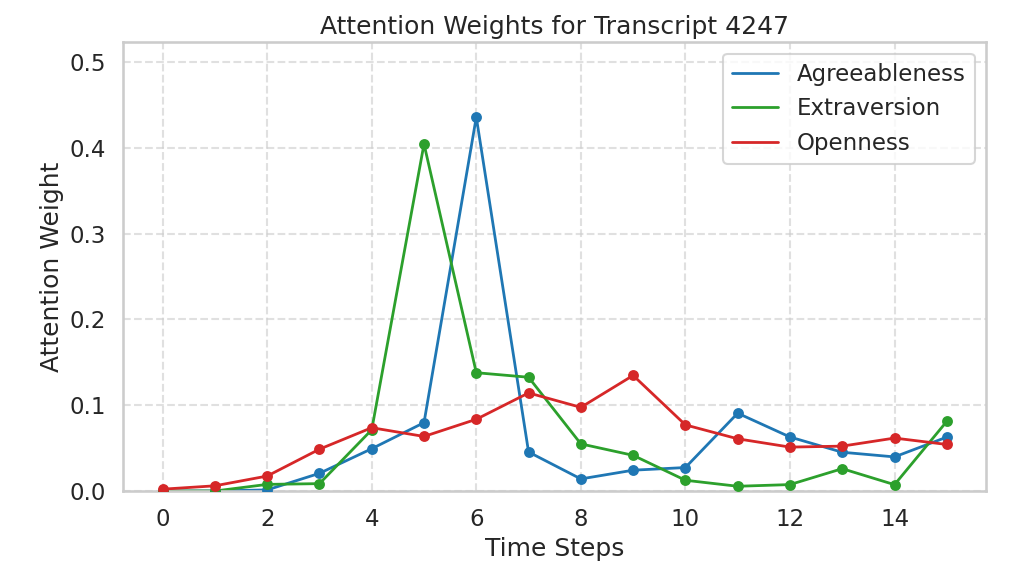}
     \caption{Context Anywhere: Attention plots of one transcript that shows a high score on openness, agreeableness, and extraversion. It shows that the context relevant to the output can be found anywhere in text.}
     \label{fig:overlap}
 \end{figure}

Using attention mechanisms, the model focused on the transcript segments most relevant to each FFM personality trait. For each trait, we randomly selected two examples from the top-10 transcripts in test set based on predicted scores and visualized the top attention windows as heatmaps in Figure~\ref{fig:heat}. For participant privacy, we list only the discussed topics in the top-5 windows rather than full text.

For openness (Fig.~\ref{fig:heat}A), high-attention windows included creative expression (photography, blogging), adventure, and excitement for career shifts in later life, in addition to other typical life challenges and struggles, in line with people who are more creative and open to experience \cite{schwaba2019structure}.

Conscientiousness examples (Fig.~\ref{fig:heat}B) emphasized career passion, responsibility, and drive, showing goal-oriented and achievement-focused individuals \cite{jackson2015conscientiousness}.

For extraversion (Fig.~\ref{fig:heat}C), participants share active, socially engaged retirements,
upbeat language about thriving through an economic downturn, and community involvement, reflecting traits of positivity and sociability \cite{wilt2015extraversion}.

Agreeableness (Fig.~\ref{fig:heat}D) examples highlighted long-term relationships, mentorship, gratitude, and emotional warmth—core features of highly agreeable individuals \cite{graziano2017agreeableness}.

Finally, neuroticism (Fig.~\ref{fig:heat}E) examples reflected emotional struggles, health-related challenges, job loss, and social isolation. These themes are commonly associated with high neuroticism \cite{widiger2017neuroticism}.

\begin{figure}
    \centering
    \includegraphics[width=7.8cm, height=5.2cm]{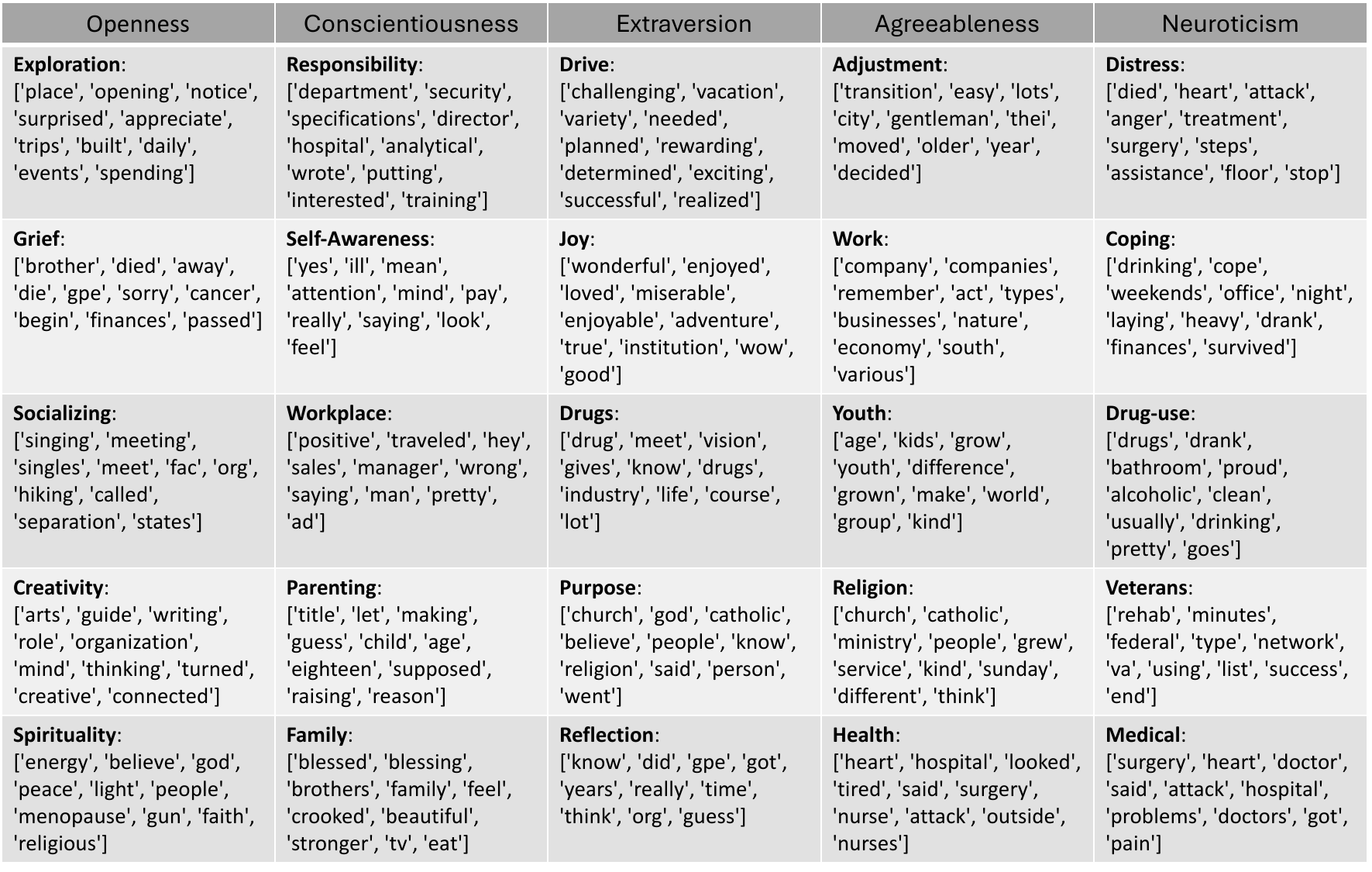}
    \caption{Top 5 correlated topics extracted from top-5 attention windows of test samples.}
    \label{fig:topics}
\end{figure}

\subsection{Impact of high attention window-text on predicted score}

We assess the impact of the most attended window-texts by removing them from selected transcripts (Figure~\ref{fig:heat}) and comparing personality predictions before and after. Across traits, removing these key segments generally led to a noticeable drop in predicted scores, emphasizing their importance.

The score drops of 6\% and 14\% for Openness, 7\% Extraversion (left), and 7\% Agreeableness (right), were observed. In these cases, attention was more evenly distributed across multiple windows, or content was repeated, reducing the impact of removing a single window.

In contrast, the score drop of 20\% and 22\% for Conscientiousness and 80\% for Agreeableness (left) were larger. These transcripts had more concentrated attention or fewer total windows, so removing a single key segment resulted in substantial information loss.

For Neuroticism, one transcript showed a 5\% increase after removal. The excluded window referenced overcoming loss, and its removal shifted focus to unresolved negative experiences, explaining the score rise. The second transcript dropped by 6\%, as the removed window conveyed emotionally significant content about disability and humility.

These patterns support the hypothesis that the attention mechanism effectively highlights critical portions of the input data, reinforcing the interpretability of the model's decision-making process.

\subsection{Context Anywhere}
In Figure \ref{fig:overlap}, we analyzed a transcript that received high predicted scores across multiple personality traits. By comparing the attention distributions for each trait, we observed that different sections of the text were emphasized, illustrating that the RNN attention can capture relevant context from any part of the text. Unlike LLMs prone to the "lost-in-the-middle" issue \citep{an2024make}, our model attends meaningfully to all segments of long texts.

\subsection{BERTopic Modelling}
To further evaluate the effectiveness of RNN attention, we extract the top 5 correlated topics for each trait using BERTopic (Figure \ref{fig:topics}) from the text of the top 5 attention windows across all test samples. Modeling details are provided in Supplementary \ref{apx:topic}. %Figure \ref{fig:topics} shows ten BERTopic clusters whose per‑participant maximum probabilities correlate most strongly with the predicted level of each FFM trait. To generate these clusters we first embedded the top three attention‑weighted transcript windows for every participant with the SentenceTransformer all‑MiniLM‑L6‑v2 model, cut the resulting vectors into overlapping sentence‑sized chunks, and reduced their dimensionality to five with UMAP using a cosine metric. We then applied HDBSCAN (minimum cluster size=2) to form 30–50 coherent topics for each trait and extracted representative keywords with c‑TF‑IDF. For every participant we took the highest probability they exhibited for each topic, assembled a topic‑by‑participant matrix, and calculated Pearson correlations between these maxima and the participant’s predicted trait scores.

Clear patterns emerge across traits. Openness is strongly linked with topics like Exploration, Socializing, Creativity, and Spirituality. Conscientiousness aligns with structured domains—reflected in terms like "department," "director," "specifications," and "training" within themes of responsibility, self-awareness, work, and family. High Extraversion corresponds with socially active and energetic terms such as “vacation,” “adventure,” and “belief” in topics related to drive, joy, and purpose. Agreeableness appears in prosocial and communal language—e.g., “gentleman,” “kids,” “nurse,” “service”—within topics like adjustment, youth, and health. In contrast, Neuroticism is linked to distress-oriented themes, such as coping, drug use, veterans’ rehab, and medical hardship.

%Several intuitive patterns emerge from these topics. Topics such as Exploration, Socializing, Creativity, and Spirituality are most strongly linked with Openness, whereas references to "department", "director", "specifications", and "training" in topics such as responsibility, self-awareness, workplace, parenting, and family correlate with Conscientiousness, reflecting orderliness and planning. Participants with high Extraversion gravitate toward topics containing social‑activity and interpersonal effectiveness terms like “vacation”, "adventure", and "belief" in drive, joy, and belief topics, respectively. For Agreeableness, the strongest topics feature prosocial and community language such as: "gentleman", "kids", "nurse", "service" in adjustment, youth, health topics, respectively.  In contrast, the topics most associated with Neuroticism center on distress, coping, drugs, veterans rehab and medical adversity.
 
These trait-topic associations, surfaced through the top-5 attention windows, highlight the meaningful alignment between language patterns and personality traits, validating both the interpretability of RNN attention and the reliability of our model.

%Together, these correlations indicate a semantically coherent theme that aligns with its respective personality trait and has been captured using top-5 attention windows showing validity of RNN attention and correctness of our model.
\section{Discussion}
%The $R^2$ values from our proposed method are higher than have been found for personality traits in prior studies \cite{park2015automatic, simchon2023online, ganesan2021empirical}. One study developed models of the FFM personality traits using Facebook status updates and found $R^2$ values ranging from 0.12 (neuroticism) to 0.18 (openness)\cite{park2015automatic}. Another using transformer modeling and again Facebook status updates found maximum $R^2$ = .12 for openness and $R^2$ = .14 for extraversion \cite{ganesan2021empirical}. Finally, another used Reddit status updates along with transformer models and found $R^2$ values ranging $.07$  to $.15$ \cite{simchon2023online}. The $R^2$ values in the present study using the RoBERTa + RNN approach are considerably higher. 

The $R^2$ values achieved by our proposed method are higher than those of other models reported in Tables \ref{tab:results} and \ref{tab:variations}, as well as the $R^2$ values reported in prior studies on personality trait prediction. For instance, \citet{park2015automatic} trained models using Facebook status updates and reported $R^2$ scores ranging from 0.12 (Neuroticism) to 0.18 (Openness). Similarly, \citet{ganesan2021empirical} employed transformer models on Facebook data and achieved a maximum $R^2$ of 0.12 for Openness and 0.14 for Extraversion. More recently, \citet{simchon2023online} used Reddit posts with transformer-based models and reported $R^2$ values ranging from 0.07 to 0.15 across traits. In contrast, our RoBERTa + RNN two-step approach yields higher $R^2$ values across all personality traits.

These findings underscore the benefits of decoupling representation learning from sequence modeling. Fine-tuning RoBERTa independently allows better adaptation to domain-specific language, producing higher-quality embeddings for downstream modeling. In contrast, end-to-end (CF) training, while more streamlined, can suffer from optimization instability, particularly when coupling PLMs with task-specific prediction heads.

While Longformer and LLaMA handle longer texts in a single pass, they involve key limitations. Longformer’s sparse attention restricts its ability to model long-range dependencies effectively. LLaMA, though capable of handling up to 128K tokens, is designed for autoregressive language modeling and is computationally intensive to fine-tune for domain-specific regression tasks. In contrast, our two-step approach using RoBERTa with lightweight sequential modeling (e.g., RNN) provides a more effective and efficient solution for long-text prediction tasks. Moreover, our attention-based interpretability analysis shows that text in high-attention windows aligns closely with personality attributes reported in the literature. This supports model validation and enabled expert review, promoting trust in clinical applications.

%Additionally, while Longformer and LLaMA enable the processing of longer texts in a single pass, they each present trade-offs. Longformer's sparse attention limits its ability to capture long-range dependencies. LLaMA, though capable of handling up to 128K tokens with architectural modifications, its decoder-only architecture is optimized for language generation tasks rather than regression. Even after adapting it with a simple regression head and applying LoRA for efficient fine-tuning, the model consistently produced negative $R^2$ scores, indicating poor alignment with the regression objective.  In comparison, our two-step framework using smaller models such as RoBERTa, followed by lightweight sequential modeling (e.g., RNN), strikes a better balance between performance and computational efficiency. This highlights the practical trade-off between model scale and architectural design when working with long-form text.

%We consistently observe that models trained with our two-step method outperform their continuous flow counterparts in terms of $R^2$, reinforcing the value of architectural modularity in long-text modeling tasks.

\section{Conclusion}

Our study demonstrates that a two-step approach: fine-tuning a compact language model followed by a lightweight contextual aggregator, can outperform larger transformer-based models in both prediction accuracy and training efficiency. While applied here to personality prediction, this interpretable, scalable method can be broadly applied to other domains requiring long-text processing using resource-efficient and interpretable models.
%The findings in this paper demonstrate the promise of language-based AI in transforming psychological assessment and personality prediction. The hierarchical model built with smaller pre-trained language models and recurrent neural network(RNN) does not increase computational costs like other Large Language Models(LLMs) with high token limits and performs better than the state-of-the-art LLMs. 

\section{Limitations}
\label{sec:limit}

While our study demonstrates the potential of language-based AI for personality assessment, several limitations should be acknowledged.

First, our dataset consists of life narrative interviews from older adults (ages 55–64) in the St. Louis area. This demographic specificity may limit the generalizability of our findings to other age groups or cultural backgrounds. Future work should validate the model on more diverse populations to ensure broader applicability. Second, our personality trait predictions rely on self-reported Five-Factor Model (FFM) scores, which are subject to biases such as social desirability and self-perception distortions. Incorporating third-party assessments or behavioral data could provide a more comprehensive evaluation of personality traits. Third, while we compare our approach with LLaMA and Longformer—two models with extended token limits—our evaluation does not include other state-of-the-art LLMs such as GPT-4 or Gemini, which may offer improved contextual understanding. Future work could explore how these models perform in personality prediction tasks and report on their interpretability and computational cost.

\bibliography{custom}

\appendix

\section{Data Preprocessing}
\label{sec:appendix}

The raw text transcripts obtained from the interviews underwent preprocessing to prepare the data for training transformer-based models. To prepare the interviews for preprocessing, raw text files were extracted from the original dataset and their corresponding interview IDs, which served as unique identifiers for each interview. The data was formatted into JSON files appropriate for training, where each entry included the complete text and corresponding FFM scores to evaluate five domain personality traits. All text for 1409 transcripts was converted to lowercase, tokenized, and stop-words were removed using the Natural Language Toolkit (NLTK) \citep{loper-bird-2002-nltk}. The cleaned text enabled clearer attribution of dialogue and enhanced readability for subsequent stages of analysis. For tokenization, we employed AutoTokenizer from the Hugging Face library to tokenize the text data.

\section{Fine-tuning details for small language models}\label{apx:fine}
We initialized pre-trained Large versions of models - RoBERTa, XLMRoBERTa, and XLNet using the Hugging Face transformers library, leveraging the pre-trained architecture to adapt to the specific characteristics of the dataset. We fine-tuned these models using the simpletransformers library for the regression task of predicting FFM personality traits. 

%All these models have 512 token limit i:e they can only process text with less than 512 tokens. 

%RoBERTa-large is a model by FacebookAI trained on 1024 V100 GPUs for 500K steps using a batch size of 8K and a sequence length of 512, and 

The models were trained with the following hyperparameters: a maximum sequence length of 512 tokens (max sequence length of the pre-trained model, a learning rate of $2e^{-5}$, a batch size of 16 for training, and 8 for evaluation. Early stopping with a patience of 5 epochs was employed to prevent overfitting, monitoring the mean absolute error (MAE) as the early stopping criterion. The model was fine-tuned over 50 epochs, with the best-performing model saved based on the lowest validation MAE. Gradient accumulation was utilized with 2 steps to optimize memory usage. 

\section{Hyperparameters for LLaMA}\label{apx:llama}
We fine-tuned the LLaMA model for 50 epochs using a batch size of 4 per device for both training and evaluation, with no gradient accumulation. Mixed precision training was enabled with FP16, while BF16 was disabled. We used a learning rate of $2e^{-4}$ with a constant scheduler and a warmup ratio of 0.03. The optimizer was set to paged\_adamw\_32bit with a weight decay of 0.001 and a maximum gradient norm of 0.3. To reduce memory usage, gradient checkpointing was disabled, and samples were grouped by length during training. Model checkpoints and logs were saved every 25 steps.

\section{Hyperparameters for Model Ablations}\label{apx:hyp}

For the continuous flow (CF) custom transformer variant (FT‑RobertaL + TF CF), the model uses 2 layers with 8 attention heads and a 2048-dimensional feed-forward network. Dropout of 0.1 is applied after the self-attention mechanism, after the feed-forward layer, and before the output layer to help mitigate overfitting. This model was trained for 50 epochs with a batch size of 32 using the Adam optimizer set to a learning rate of $1e^{-4}$.

In the two‑step transformer $(FT‑RobertaL +TF 2‑step)$ approach, the custom architecture remains similar in terms of attention heads, layers, and dropout application; however, a different training regime is employed. Here, training is carried out with a per‑device batch size of 8, a learning rate of $5e^{-5}$, and weight decay of 0.01. Dropout is consistently applied after both the self-attention and feed‑forward layers to further prevent overfitting. The two step approach uses the same numeric embeddings that were used for the FT-RobertaL+RNN experiment.

For the RNN-based continuous flow $(FT‑RobertaL +RNN CF)$, there are 2 layers of GRU units to capture sequential dynamics. This variant adopts the same training duration (50 epochs), batch size (32), and Adam optimizer with a $1e^{-4}$ learning rate as its transformer counterparts. In this RNN setup, dropout of 0.1 is additionally applied in the fully connected regression head to ensure regularization despite the more sequential nature of the model.

\vspace{1em}

%\section{RNN Mathematical Notation}\label{apx:rnnmat}
% Here we explain the mathematical notation for the FT-RobertaL+RNN experiment. 

% \begingroup\scriptsize
% \begin{equation}
% (w_1,\dots,w_T) \;=\;
% \operatorname{CLS}\!\bigl(\texttt{RoBERTa}(\text{SlidingWindow}(x))\bigr)
% \end{equation}
% \endgroup  

% \noindent{}
% The raw text $x$ is split into $T$ overlapping windows.  
% Each window is fed through RoBERTa, and we keep the model’s \texttt{[CLS]}
% embedding.  The result is a sequence of fixed‑length vectors
% $(w_1,\dots,w_T)$ basically one for every window.

% \vspace{1em}

% \begingroup\footnotesize
% \begin{equation}
% (a_1,\dots,a_T) \;=\;
% \operatorname{softmax}\!\bigl(\operatorname{score}(w_1,\dots,w_T)\bigr)
% \end{equation}
% \endgroup

% \noindent\textbf{}
% Each hidden vector $w_i$ is passed through a tiny feed‑forward
% \textit{score} function which maps each timestep and the resulting scores are normalized and converted to attention
% weights with \texttt{softmax}, guaranteeing all $a_i$ are non‑negative and
% sum to 1.

% \vspace{1em}

% \begin{equation}\label{eq:rnn3}
% \hat{y} \;=\;
% w_o^{\mathsf T}\!\left(\sum_{i=1}^{T} a_i\,w_i\right) + b_o
% \end{equation}

% \noindent{}
% A weighted average of the hidden vectors (using the attention weights
% $a_i$) forms a single context vector.  A final linear layer
% with parameters of $(w_o,b_o)$ maps that vector to the scalar prediction
% $\hat{y}$.

\section{Topic Modeling}\label{apx:topic}
To extract latent topics from a collection of text segments from top 5 high-attention windows. We concatenated all 5 high-attention window text. We employed the BERTopic framework with several customized components and hyperparameters. Semantic embeddings were first generated using the pre-trained SentenceTransformer model "all-MiniLM-L6-v2", which converts each utterance into a dense vector representation. To reduce noise from filler words common in spoken language, a custom stop word list was created by augmenting ENGLISH\_STOP\_WORDS with terms such as "um", "uh", "like", and "okay". A CountVectorizer was then initialized with this stop word list and configured with min\_df set to max(1, int(len(segment) * 0.001)) to ignore extremely rare terms, and max\_df=0.99 to exclude overly frequent terms. Dimensionality reduction was performed using UMAP with hyperparameters n\_neighbors=10, n\_components=5, min\_dist=0.0, and metric="cosine", while clustering was conducted using HDBSCAN with min\_cluster\_size=5, metric="euclidean", and prediction\_data=True to enable probabilistic topic assignments. These components were integrated into BERTopic with language="english", calculate\_probabilities=True, and nr\_topics="auto" to allow automatic topic number determination. The resulting model was then fit to the input utterances, producing topic labels and probability distributions for each text segment. After finding topics we calculated the probability of each topic in the concatenated text from top 5 high-attention windows and used these probabilities to find Pearson correlation between personality score and topic probabilities.

\section{Bias Analysis}\label{apx:bias}
%In this analysis, we analyzed if there is any bias in our model between predicted values and the length of the transcripts.

In order to investigate the potential bias of the model performance with respect to the length of input transcripts, the correlation of the length of transcripts with the predicted values was examined. The p-value for the correlation of -0.05, 0.04, 0.01, 0.03, and -0.05 for O, C, E, A, N traits between the length of the interview text and predicted values is more than 0.05. The average p-value is 0.5. This would suggest that none of the models have a statistically significant relation between the transcript lengths and predicted values. We also determine the Pearson correlation between gender and the predicted output for each model was conducted. The findings, show that the relationship between gender and expected production was not statistically significant in four of the five models (C, E, A, and N). However, a weak but statistically significant correlation was observed in the Model for Openness (r = 0.1343, p = 0.0276), indicating a slight bias associated with this trait and gender.

\end{document}